\documentclass[10pt,twocolumn,letterpaper]{article}

\usepackage{wacv}
\usepackage{times}
\usepackage{epsfig}
\usepackage{graphicx}
\usepackage{amsmath}
\usepackage{amssymb}

\usepackage{comment}
\usepackage{color}
\usepackage{tabulary}
\usepackage{epsfig}
\usepackage{graphics}
\usepackage{mathtools}
\usepackage{xcolor}
\usepackage{multirow}
\usepackage{hhline}
\usepackage{makecell}
\usepackage{lineno}
\usepackage{booktabs}
\usepackage{url} 
\usepackage{array, boldline, rotating}

\wacvfinalcopy 

\ifwacvfinal
\def\assignedStartPage{1} 
\fi

\ifwacvfinal
\usepackage[breaklinks=true,bookmarks=false]{hyperref}
\else
\usepackage[pagebackref=true,breaklinks=true,colorlinks,bookmarks=false]{hyperref}
\fi

\ifwacvfinal
\else
\fi

\begin{document}

\title{Resisting Crowd Occlusion and Hard Negatives \\for Pedestrian Detection in the Wild}

\author{Zhe Wang\\
Beihang University\\
{\tt\small wangzhewz@buaa.edu.cn}
\and
Jun Wang\\
Beihang University\\
{\tt\small wangj203@buaa.edu.cn}
\and
Yezhou Yang\\
Arizona State University\\
{\tt\small yz.yang@asu.edu}
}

\maketitle

\begin{abstract}
   Pedestrian detection has been heavily studied in the last decade due to its wide application. Despite incremental progress, crowd occlusion and hard negatives are still challenging current state-of-the-art pedestrian detectors. In this paper, we offer two approaches based on the general region-based detection framework to tackle these challenges. Specifically, to address the occlusion, we design a novel coulomb loss as a regulator on bounding box regression, in which proposals are attracted by their target instance and repelled by the adjacent non-target instances. For hard negatives, we propose an efficient semantic-driven strategy for selecting anchor locations, which can sample informative negative examples at training phase for classification refinement. It is worth noting that these methods can also be applied to general object detection domain, and trainable in an end-to-end manner. We achieves consistently high performance on the Caltech-USA and CityPersons benchmarks\footnote{Code will be publicly available upon publication.}.
\end{abstract}

\section{Introduction}\label{sec:intro}
Pedestrian detection is a important research topic in computer vision and has attracted massive research interest in recent years~\cite{benenson2014ten,dollar2009integral,dollar2009pedestrian,mao2017can,zhang2016faster,zhang2016far}. It aims to predict accurate bounding boxes enclosing each pedestrian instance and serves as a key component of various real-world applications such as autonomous driving, robotics, and intelligent video surveillance.

Although promising results have been achieved, crowd occlusion and hard negatives are still remain as two greatest challenges in this domain. Since most pedestrian detectors adopt region-based framework, we systematically study the impact of crowd occlusion and hard negatives on region localization and classification, respectively.

\begin{figure}[t]
\includegraphics[width=1.0\linewidth]{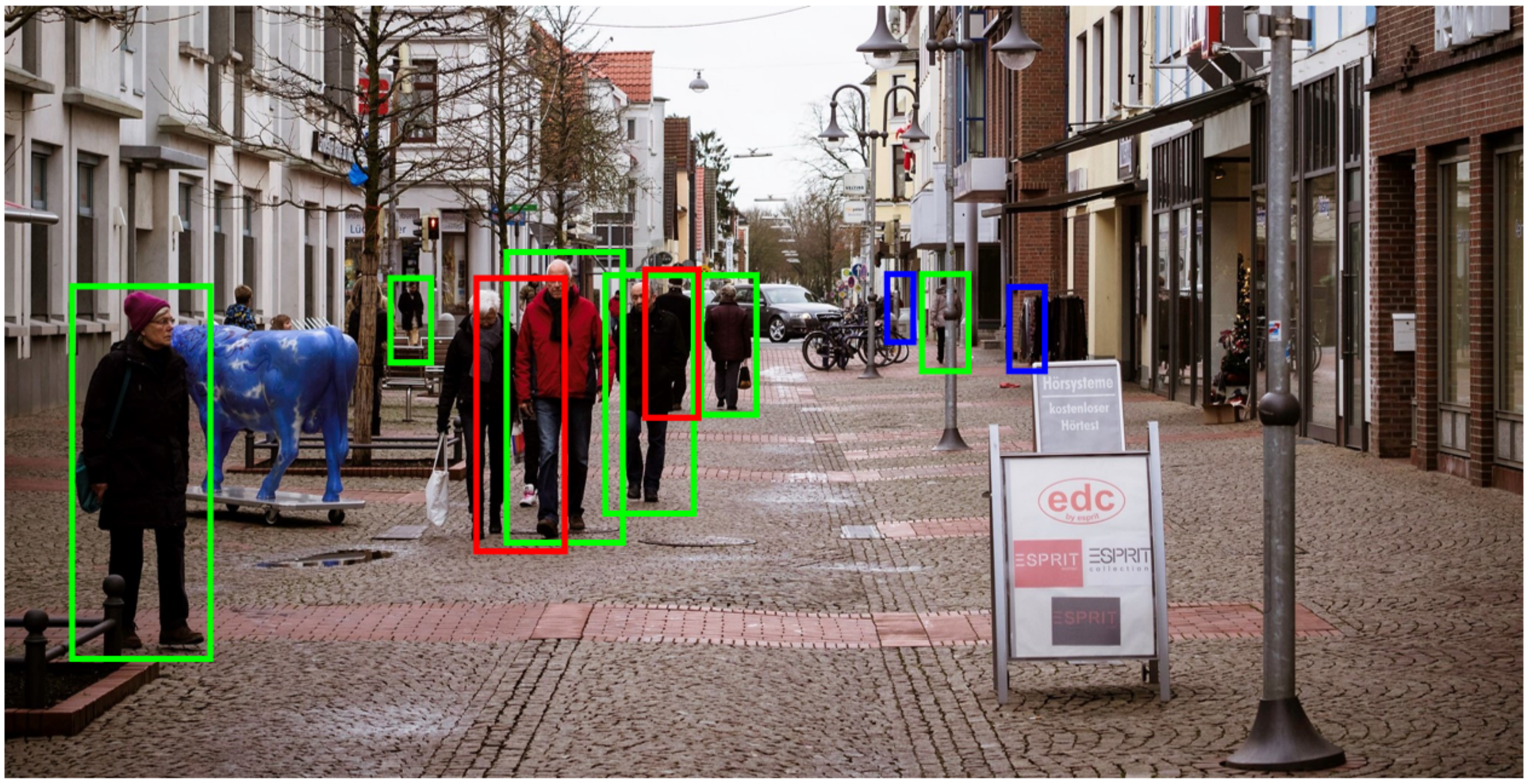}
\caption{Pedestrian detection in the wild. Green boxes represent correct predictions. Red boxes indicate missed targets and misplaced prediction caused by the occlusion. Blue boxes are the false positive predictions caused by the hard negatives.}
\label{fig:overview}
\end{figure}

Crowd occlusion (also known as \emph{intra-class} occlusion) is the most significant barrier for accurate pedestrian detection in the wild and also the major occlusion case in most pedestrian datasets. When pedestrians gather together and occlude each other, detector is prone to be disturbed by the instance that is adjacent to the target and generates bounding boxes among their overlaps (as the left red box in Figure~\ref{fig:overview}). Even worse, during non-maximum suppression (NMS) processing, misplaced boxes with higher confidence scores may suppress the accurate ones or bigger boxes may suppress their neighbouring small ones. At the same time, it also makes detector sensitive to the threshold of NMS, as a higher threshold brings in more false positives while a lower threshold leads to more missed detections~\cite{wang2018repulsion}.
Hard negatives usually share similar shape with human body (e.g. pillars, light poles). These objects frequently appears in the common scenarios of real world and we define them as human-like structures in this paper. Due to the complicated light condition and variant resolutions, detector is unable to correctly recognize these objects and may assign them with higher probabilities to person rather than background (as the blue boxes in Figure~\ref{fig:overview}).

Several efforts have been made to tackle these two challenges. For the former one, previous methods like~\cite{wang2018repulsion,zhang2018occlusion} introduce an extra penalty term on the bounding box regressor to constrain each sampled proposals. But their regularization is incomplete and is likely to get conflict with the original regression function.
Others like~\cite{Bodla_2017_ICCV,Liu_2019_CVPR} try to mitigate negative impact by refining traditional greedy-NMS~\cite{girshick2014rich}. However, the effectiveness of these post-processing methods are restricted by the accuracy of the predictions. 
For the latter one, several hard/soft-sampling methods~\cite{Lin_2017_ICCV,shrivastava2016training} are proposed to either mine hard negatives or re-weight each sample to train the classifier. Both methods are loss-driven, which means they may easily neglect the semantic relations of the backgrounds with the foregrounds, that can be useful clues for negative example mining. 

In this paper, we put forward two novel approaches that based on Faster R-CNN framework~\cite{ren2015faster} to tackle the aforementioned challenges. For crowd occlusion, the key point is to generate accurate bounding boxes in occluded scenes. Inspired by the Coulomb Force~\cite{halliday2013fundamentals} between two electric charges, we define \emph{Attractive Force} between proposal and its target ground truth as well as \emph{Repulsive Force} between proposal and its non-target ground truth. With this insight, we build a \textbf{physics} modeling and use the energy consumption, calculated by \textbf{work formula}, as the measurement of loss value. The new loss function, termed as \emph{Coulomb Loss} (CouLoss), works as a regulator that constrains each proposal during regression process.
As for hard negatives, the breakthrough could be achieved by avoiding misclassification of human-like structures. To this end, we propose an efficient anchor location selecting strategy functioning as \emph{informative} negative examples mining. By adding an extra branch on region proposal network (RPN)~\cite{ren2015faster}, a probability map is yielded and we only process the anchors whose probabilities are over a dynamic threshold. These \emph{informative} negative examples not only cause high loss values but also have semantic relations with pedestrian foregrounds.

To validate the effectiveness of these improvement, we conduct extensive experiments on both Caltech-USA~\cite{dollar2009pedestrian} and CityPersons~\cite{zhang2017citypersons} benchmarking datasets. The main contributions are as follows:

\begin{itemize}
\setlength{\itemsep}{0pt}
\setlength{\parskip}{0pt}
\setlength{\parsep}{0pt}
\item For crowd occlusion resolving, we design a new CouLoss on the basis of \textbf{work formula} that serves as a regulator for bounding box regression. It enforces proposals to locate compactly around their targets, meanwhile, penalizes proposals for shifting to other non-targets.
\item For hard negatives handling, we modify RPN~\cite{ren2015faster} with an extra branch for pedestrian existence prediction, and propose a novel sampling method to capture informative negative examples to train the classifier.
\item Experimental results show the superiority of the proposed methods on pedestrian detection benchmarks. We also carry out experiments on PASCAL VOC dataset~\cite{everingham2010pascal} to validate that our approaches are applicable for other general object detection tasks.
\end{itemize}


\section{Related Work}
We briefly review recent work on CNN-based pedestrian detector and discuss related researches on the two target challenges: crowd occlusion and hard negatives.

\noindent\textbf{CNN-Based Pedestrian Detectors.} Recently, CNN-based methods have dominated the field of pedestrian detection~\cite{brazil2017illuminating,lin2018graininess,mao2017can,noh2018improving,tian2015deep,zhang2018occluded,zhou2018bi} and achieved state-of-the-art performance~\cite{wang2018repulsion,liu2018high,zhang2018occlusion} on several benchmarks: INRIA~\cite{dalal2005histograms}, ETH~\cite{ess2008mobile}, Caltech-USA~\cite{dollar2009pedestrian}, and CityPersons~\cite{zhang2017citypersons}. Most of these models adopt region-based approaches where detectors are trained to localize and classify sampled regions. Similar to general object detection, there are also two different frameworks in pedestrian detection. Two-stage framework like~\cite{wang2018repulsion,zhang2017citypersons,zhang2018occlusion} first generate a set of candidate proposals and then sample a small batch of proposals for further bounding box regression and classification. One-stage frameworks like~\cite{Liu_2018_ECCV,noh2018improving,Song_2018_ECCV} directly predict bounding box offsets and class scores from all anchors at each coordinate.

\noindent\textbf{Crowd occlusion resolving.} Attention models are proposed to improve the feature representation of visible parts.~\cite{lin2018graininess} generates scale-aware attention masks in semantic segmentation manner.~\cite{zhang2018occluded} employs a channel-wise attention mechanism from three different attention modules. Anchor-free methods are used to directly predict bounding boxes of each target. In~\cite{huang2015densebox}, bounding boxes are learned through a single convolutional neural network.~\cite{liu2018high} predicts the center points of targets and regress the height and width of them. Other solutions formulate the issue as a regression problem. To better allocate proposals to each pedestrian,~\cite{wang2018repulsion} proposes \emph{Repulsion Loss} to keep proposals away from the non-targeted ground truth and their proposals, while~\cite{zhang2018occlusion} comes up with \emph{Aggregation Loss} that enforces proposals to locate compactly around each other when they belong to the same target.

Our method shares a common spirit with~\cite{wang2018repulsion,zhang2018occlusion} where an extra regulation term is used in loss function to guide proposal regression. While the distinctive part is that we simultaneously consider both attraction and repulsion progress in the extra regulation term, which makes our constraints theoretically more complete than~\cite{wang2018repulsion,zhang2018occlusion}. What's more, we propose a \textbf{physics} framework to unify these two progresses and make them compatible with each other.

\noindent\textbf{Hard negatives handling.} Multi-classifier is a common structure to refine classification results.~\cite{tian2015deep} employs different patterns that can generate a pool of parts for classifier to choose.~\cite{du2017fused} trains multiple classifiers in parallel phase and fuse the scores to filter candidates. A set of grid score map from multi-stage is generated by~\cite{noh2018improving} to revise final prediction scores. Methods like~\cite{li2019gradient,Lin_2017_ICCV,pang2019libra,shrivastava2016training} balance the region of interest (ROI) to train the classifier.~\cite{shrivastava2016training} proposes a hard example mining method which only samples negative proposals with high loss values.~\cite{Lin_2017_ICCV} designs Focal Loss which assigns different weights to all proposals based on their probabilities.

We believe the problem is caused by under-sampling of useful negative examples (foreground-background imbalance). Our strategy is mining human-like structures as negative examples to train the classifier. Different from current sampling methods like~\cite{li2019gradient,Lin_2017_ICCV,shrivastava2016training} which are loss-driven and~\cite{cao2019prime,pang2019libra} which are IoU-driven, our method samples regions that have high semantic relativity with pedestrians. We term these regions as \emph{informative} negative examples since they have higher probabilities to contain human-like structures than others.


\section{Proposed Approach} \label{sec:approach}
In this section, we systematically analyze the aforementioned critical issues and then offer our solutions. Besides the benefit to the performance, an important advantage of our methods is that we do not increase any computational cost during inference phase.

Specifically, we introduce coulomb loss which is especially designed for crowd scenes in Section~\ref{sec:coulombloss}. Then, a novel anchor sampling method is proposed in Section~\ref{sec:anchorloc} to mine informative negative examples. Finally, in Section~\ref{sec:architecture}, we present the network architecture and the loss function for end-to-end training.

\subsection{Coulomb Loss} \label{sec:coulombloss}
Our key idea for resolving crowd occlusion is to regularize the bounding box regression with extra constraints. Inspired by the physical property of electric charge, we regard each bounding box as a single charge. Then, we define \emph{Attractive Force} $F_a$ and \emph{Repulsive Force} $F_r$ as the interaction between a proposal and its target/non-target ground truth, respectively. Suppose $\mathcal{P_+}$ is the set of proposals that has high Intersect over Union (IoU) value (\emph{e.g.}, $IoU>0.5$) with ground truth. We set proposals $P_p,P_n\in\mathcal{P_+}$, and $G_i$, $G_j$ are the target ground truth of $P_p$ and $P_n$, respectively. For the convenience of analysis, we form a triplet $\big<G_i,P_p,P_n\big>$, where $G_i$ is the target while $P_p$ and $P_n$ are the belonging positive and negative sample.

In physics, \textbf{work}\footnote{\url{https://en.wikipedia.org/wiki/Work_(physics)}} is used to measure the energy consumption for moving an object from one place to another. Rationally, we can set this value as the cost of pulling $P_p$ toward $G_i$ or pushing $P_n$ away from $G_i$, which is exactly the loss value we need. To utilize the \textbf{work formula} $W=F\cos{\theta}\cdot s$ for calculating, we need to build a physics modeling at bounding box level.

First and the most important, we have to define the \emph{Force} between boxes which is related to their distance. Since IoU is a widely used metric for measuring the closeness between two bounding boxes, we refer to the objective function of IoU Loss~\cite{yu2016unitbox} and formulate the forces as:
{
\begin{equation}
\begin{split}
F_a&=-\mathrm{ln}(IoU(G_i, P_p)) \\
F_r&=-\mathrm{ln}(1-IoU(G_i, P_n)).
\end{split}
\label{eq:forcefunc}
\end{equation}
}
Note that the forces only exist when there is an overlap between proposal and ground truth (\emph{i.e.}, $IoU>0$). From Eq.~\ref{eq:forcefunc} we can see that the lower closeness between a proposal and its target instance, the stronger \emph{Attractive Force} will be applied to the proposal, whilst the higher closeness between a proposal and its non-target instance, the stronger \emph{Repulsive Force} will be applied. Numerically, $-\mathrm{ln}(x)$ gets extreme large when $x$ approaches to 0, which will make the training process unstable. As in~\cite{Tian_2019_ICCV}, we only select proposals whose center points fall into the region of their corresponding ground truth boxes.

\begin{figure}[t]
\includegraphics[width=1.0\linewidth]{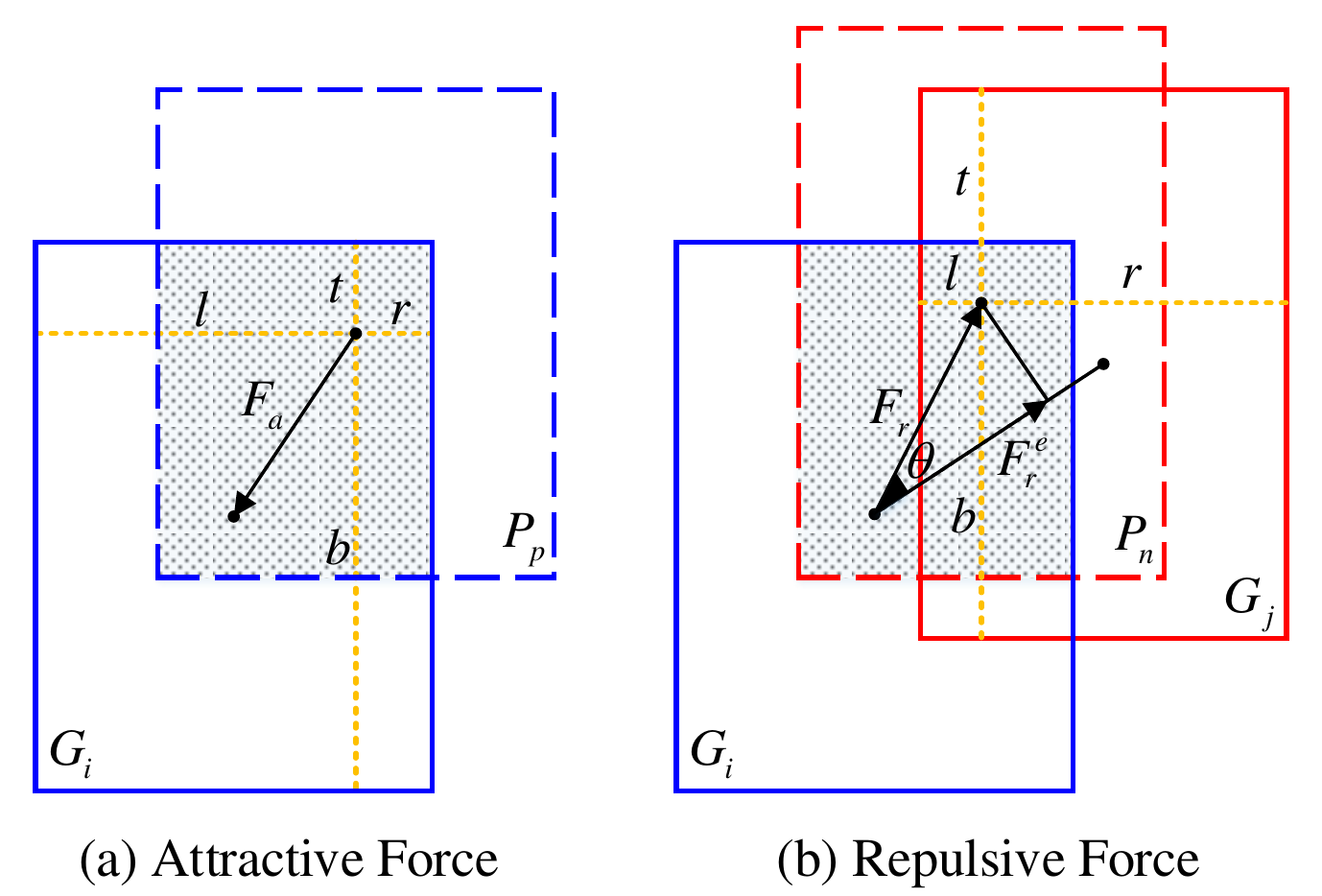}
\caption{The Attractive and Repulsive Forces between proposal and ground truth. The direction of Attractive Force is always toward the target, while the direction of Repulsive Force may deviate from the target.}
\label{fig:force}
\end{figure}

\begin{figure*}[t]
\centering
\includegraphics[width=1.0\linewidth]{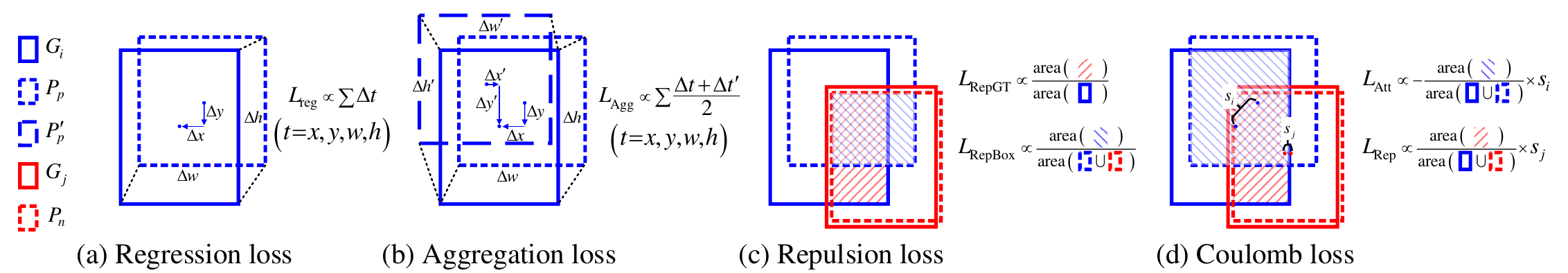}
\caption{Illustrations of bounding box regression loss, as well as the attraction and repulsion terms in different objective functions.}
\label{fig:compare}
\end{figure*}

Originally, $\cos{\theta}$ is introduced because the force may not always has the right direction that moving object toward its target location. In this case, only part of the force is effective. It is also reasonable to follow the same definition in box regression, as illustrated in Figure~\ref{fig:force}. The \emph{Attractive Force} always pulls $P_p$ at the desired direction, but the \emph{Repulsive Force} may push $P_n$ deviated from its original target when its direction is not on the center line of $(G_i,G_j)$. To handle such situation, we introduce Effective Force ($F^e$) as the component of original force:
{
\begin{equation}
\begin{split}
F_{a}^{e} &= F_{a}\cos \theta, \quad \theta = 0 \\
F_{r}^{e} &= F_{r}\cos \theta, \quad \theta = \angle P_{n}G_{i}G_{j},
\end{split}
\label{eq:effectiveforce}
\end{equation}
}
where $\cos{\theta}$ can be calculated by \textbf{the law of cosines} since we have the coordinates of each proposal and ground truth. With Eq.~\ref{eq:effectiveforce}, $F_a^e$ is defined as the force pulling $P_p$ toward $G_i$, and $F_r^e$ is the force pushing $P_n$ toward $G_j$.

At last, we define $s$ as the geometry distance between proposal and its target ground truth:
{
\begin{equation}
s = \sqrt{(1-\frac{\mathrm{min}(l,r)}{W_G/2})\cdot(1-\frac{\mathrm{min}(t,b)}{H_G/2})}.
\label{eq:distance}
\end{equation}
}
In Eq.~\ref{eq:distance}, $l,r,t,b$ are the distance of the center point of proposal to the left, right, top, bottom border of its ground truth respectively, as shown in Figure~\ref{fig:force}.

The work values and overall CouLoss are calculated as:
{
\begin{gather}
W_{a}^{<G_i,P_p,P_n>} = F_a\cos{\theta_a}\cdot s_{<G_i,P_p>} \nonumber \\
W_{r}^{<G_i,P_p,P_n>} = F_r\cos{\theta_r}\cdot s_{<G_i,P_n>} \label{eq:workvalue} \\ 
L_{cou} = \frac{1}{|\mathcal{G}|}\sum_{G \in \mathcal{G}}\sum_{P_p,P_n \in \mathcal{P_+}}(W_{a}^{<G_i,P_p,P_n>} + W_{r}^{<G_i,P_p,P_n>}). \nonumber
\end{gather}
}
It is worth noting that we ignore the cases when $W^{triplet} \leqslant 0$ in Eq.~\ref{eq:workvalue} since they do not make any work that move proposals toward their target locations. Last but not least, this new CouLoss can benefit both RPN and Fast-RCNN~\cite{girshick2015fast} modules in Faster R-CNN framework.

The illustrations of attraction and repulsion terms in CouLoss and other related objection functions are shown in Figure~\ref{fig:compare}. Specifically, we use the diagrams of attractive constraint on $P_p$ and repulsive constraint on $P_n$ to visualize the differences from AggLoss~\cite{zhang2018occlusion} and RepLoss~\cite{wang2018repulsion}. Firstly, instead of including interaction of proposal-proposal, CouLoss only considers the interaction of proposal-gt. This is to avoid conflict with the original box regression loss. Also, the inaccurate locations of the proposals during training will misguide the proposal-proposal modeling process. Secondly, CouLoss restricts the repulsion in the crowd scenes where there is an intrinsic overlap between targets $G_i$ and $G_j$. In the cases of Figure~\ref{fig:compare}(c) and (d), the repulsion existed in RepLoss will push the well-regressed $P_n$ away from $G_j$. While in CouLoss, the Repulsive Force diminishes to 0 when $P_n$ is close to $G_j$.

\subsection{Anchor Location Selecting} \label{sec:anchorloc}
Human-like structures always act as hard negatives in pedestrian detection due to the foreground-background class imbalance. This problem is mainly caused by the sampling method in detection framework. For instance, in RPN~\cite{ren2015faster}, since the only sampling principle for negative examples is the IoU with ground truth bounding boxes (\eg, $IoU<0.3$), there is a high probability for negative proposals to be sampled in easily distinguished area (\eg, sky and road). Classifier trained with these negative examples will soon converge and lose the ability to learn hard ones. To this end, our solution is trying to mine informative negative examples to train the classifier. 

\begin{figure}[t]
\includegraphics[width=1.0\linewidth]{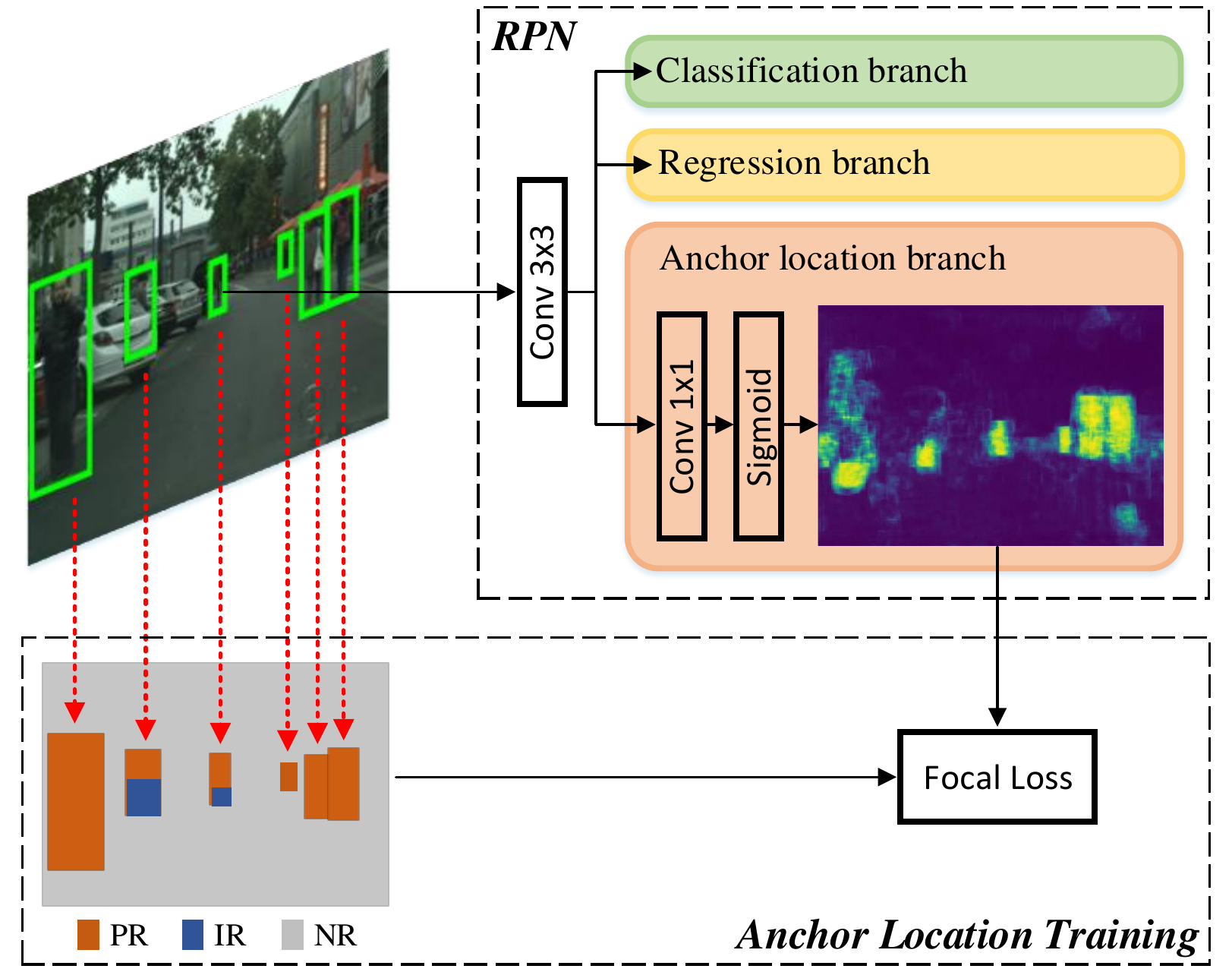}
\caption{The extra anchor location branch on the RPN module. The new added branch yields a probability map of the existence of pedestrian. During training, a dynamic threshold is used to filter out low-probability regions, and the model only select anchors whose centers fall into the remained regions. To generate the anchor location targets, we use both full body and visible body box annotation to define $PR$, $IR$, and $NR$.}
\label{fig:rpn}
\end{figure}

To better sample informative negative examples, we put forward a novel scheme that can erase anchors from easily distinguished areas. As shown in Figure~\ref{fig:rpn}, an anchor location branch is added on RPN module which can yield a probability map representing the existence of pedestrian at each coordinate. Based on the fact that human-like structures usually have similar feature representations with humans, the high value regions on the probability map are also more likely to contain hard negatives. Therefore, instead of uniformly matching every anchors on the full-scale feature map, we only select the anchors whose center points fall into the regions that have larger probabilities than a threshold during the training phase. We set the root mean square value of the probability map as the dynamic threshold ($\epsilon_a$) which can adaptively adjust based on the input images.

To train the anchor location branch, we employ the ground-truth bounding boxes to generate a binary score map where $1$ indicates selected location and $0$ indicates the rest. In specific, we categorize three types of regions on each score map as shown in Figure~\ref{fig:rpn}.

(1) Positive region ($PR$). We define the areas of visible bounding boxes $(x_v,y_v,w_v,h_v)$ as $PR$, because these parts provide the most valuable semantic information.

(2) Ignored region ($IR$). The non-visible part is generated by excluding visible part from full-body bounding boxes $(x_f,y_f,w_f,h_f)$. We mark this area $(x_f,y_f,w_f,h_f)-(x_v,y_v,w_v,h_v)$ as ignored region. These regions are harmful to classifier, because proposals in $IR$ might be labeled as positive but without any human feature representations (see Figure 4(b) in~\cite{zhou2018bi} for further details).

(3) Negative region ($NR$). The rest part of the score map only contains background information and is regarded as $NR$.

The proposed anchor selecting strategy can rapidly narrows down the searching space of initialized anchor ($\emph{e.g.}\sim100k$) to a small scale ($\emph{e.g.}\sim10k$). As shown in Figure~\ref{fig:anchor}, it effectively filter out the low-probability regions and select the anchors that have strong semantic relations with foregrounds (\eg human-like structures) as negative examples. Please note that we only use this strategy during training phase and we can drop the anchor location branch for computational cost saving during inference phase. It is also worth noting that, though our modification to RPN is similar to the changes in~\cite{wang2019region}, these two models share completely different designing goals.~\cite{wang2019region} is trying to generate accurate bounding boxes for foregrounds by learnable shape and location, while we propose to sample negative proposals that have high scores on location confidence map. Our setting is based on the fact that human-like structures have strong semantic relations with humans, and we make use of these relations as the clue to mine informative negative examples.

\subsection{Network Architecture} \label{sec:architecture}
We implement the two proposed methods on the widely used Faster R-CNN framework~\cite{mmdetection} and uses VGG-16~\cite{simonyan2014very} as the backbone. To better fulfill pedestrian detection task, the detector is modified following the settings in~\cite{zhang2017citypersons}. 

The final loss function is jointly optimized with the following losses:
{
\begin{equation}
L = L_{ori} + \alpha L_{cou}^{rpn} + \beta L_{cou}^{rcnn} + \gamma L_{loc},
\label{eq:lossfunction}
\end{equation}
}
where $L_{ori}$ represents the original classification and regression loss in both RPN and Fast-RCNN modules. $L_{cou}^{rpn}$ and $L_{cou}^{rcnn}$ are the extra regularization terms for regression, and $L_{loc}$ is the Focal Loss~\cite{lin2017focal} for training binary classification for anchor location.
Coefficients $\alpha$, $\beta$, and $\gamma$ are the hyperparameters used to balance auxiliary losses.


\section{Experiments}
\subsection{Experimental Setting}
\textbf{Datasets.} We conduct experiments on two benchmarks: Caltech-USA~\cite{dollar2009pedestrian} and CityPersons~\cite{zhang2017citypersons}. Both benchmarks contain annotations for the visible areas. We use Caltech-USA10x which samples 42,782 frames and 4,024 frames as training and testing datasets respectively. The refined annotation provided by Zhang \etal~\cite{zhang2016far} is used in related experiments. CityPersons is a more challenging dataset derived from Cityscapes~\cite{cordts2016cityscapes}. It includes 5,000 images in total and 2,975, 500, 1,525 images for training, validation and testing respectively.

\noindent\textbf{Implementation details.} As a common convention, we horizontally flip training images for pre-processing. The Adam solver with 0.0001 weight decay is adopted to optimize the network on 1 Nvidia TITAN GPU. A mini-batch involves 2 image per GPU for computational resource constraint. We set the base learning rate to 0.0001 and train the network for 16 epochs and 12 epochs on Caltech-USA and CityPersons respectively. Hyperparameters $\alpha$, $\beta$, and $\gamma$ are empirically set to 1.

\noindent\textbf{Evaluation protocols.} The models are evaluated by log-average miss rate ($\rm MR^{-2}$), which is the average value over the false positive per image (FPPI) range of $[10^{-2}, 10^0]$. The lower value represents better pedestrian detection performance.
To further evaluate performances in occluded scenes, we also present the scores on different subsets as introduced in~\cite{dollar2009pedestrian,zhang2017citypersons}.

\begin{table*}[t]
\begin{center}
\setlength{\tabcolsep}{3.5pt}
\begin{tabular}{p{1.0cm}<{\centering}|p{2.0cm}<{\centering}|p{2.0cm}<{\centering}|p{1.0cm}<{\centering}|c|p{1.0cm}<{\centering}p{1.0cm}<{\centering}p{1.0cm}<{\centering}}
\toprule[1pt]
\multicolumn{2}{c|}{Method} &Framework &Scale &{\em Reasonable} &{\em Heavy} &{\em Partial} &{\em Bare} \\
\hline
\multicolumn{2}{c|}{ATT-part~\cite{zhang2018occluded}} & VGG-16 &$\times1$ &16.0 &56.7 &- &- \\
\multicolumn{2}{c|}{TLL~\cite{song2018small}} & ResNet-50 &$\times1$ &15.5 &53.6 &17.2 &10.0 \\
\multicolumn{2}{c|}{FRCNN~\cite{zhang2017citypersons}} & VGG-16 &$\times1$ &12.9 &50.5 &- &- \\
\multicolumn{2}{c|}{ALFNet~\cite{Liu_2018_ECCV}} & ResNet-50 &$\times1$ &12.0 &51.9 &11.4 &8.4 \\
\multicolumn{2}{c|}{RepLoss~\cite{wang2018repulsion}} & ResNet-50 &$\times1.3$ &11.6 &55.3 &14.8 &7.0 \\
\multicolumn{2}{c|}{MGAN~\cite{Pang_2019_ICCV}} & VGG-16 &$\times1$ &11.5 &51.7 &- &- \\
\multicolumn{2}{c|}{Bi-Box~\cite{zhou2018bi}} & VGG-16 &$\times1.3$ &11.2 &\textcolor{red}{44.2} &- &- \\
\multicolumn{2}{c|}{OR-CNN~\cite{zhang2018occlusion}} & VGG-16 &$\times1.3$ &11.0 &51.3 &13.7 &5.9 \\
\multicolumn{2}{c|}{CSP~\cite{liu2018high}} & ResNet-50 &$\times1$ &11.0 &49.3 &\textcolor{red}{10.4} &7.3 \\
\hline
\hline
\multirow{4}{*}{Ours} &Baseline &\multirow{4}{*}{VGG-16} &\multirow{4}{*}{$\times1.3$} &12.7 &54.2 &14.4 &7.3  \\
&+ CouLoss & & &\textcolor{blue}{10.5} &51.9 &11.3 &\textcolor{red}{5.8}  \\
&+ ALS & & &11.1 &51.2 &12.2 &\textcolor{blue}{5.9}  \\
&+ both & & &\textcolor{red}{10.4} &\textcolor{blue}{46.9} &\textcolor{blue}{10.7} &\textcolor{red}{5.8}  \\
\bottomrule[1pt]
\end{tabular}
\end{center}
\caption{Pedestrian detection results on CityPersons validation set. ALS is the short form of Anchor Location Selecting. All models are trained on the trainset. We use $\rm MR^{-2}$ as the performance comparing each detectors. The best and the second best are highlighted in red and blue.}
\label{tab:cityperson}
\end{table*}

\begin{table}[t]
\begin{center}
\setlength{\tabcolsep}{7.5pt}
\begin{tabular}{c|c p{1.1cm}<{\centering}p{1.1cm}<{\centering}}
\toprule[1pt]
Method &{\em Reasonable} &{\em Heavy} &{\em All} \\
\hline
DeepParts~\cite{tian2015deep} &11.9 &60.4 &64.8 \\
ATT-part~\cite{zhang2018occluded} &10.3 &45.2 &\textcolor{red}{54.5} \\
MS-CNN~\cite{cai2016unified} &10.0 &59.9 &60.9 \\ 
RPN+BF~\cite{zhang2016faster} &9.6 &74.4 &64.7 \\
Bi-Box~\cite{zhou2018bi} &7.6 &\textcolor{red}{44.4} &- \\
SDS-RCNN~\cite{brazil2017illuminating} &7.4 &58.6 &61.5 \\
RepLoss~\cite{wang2018repulsion} &5.0 &47.9 &59.0 \\
CSP~\cite{liu2018high} &\textcolor{red}{4.5} &45.8 &\textcolor{blue}{56.9} \\
\hline
\hline
Ours &\textcolor{blue}{4.9} &\textcolor{blue}{45.5} &57.0 \\
\bottomrule[1pt]
\end{tabular}
\end{center}
\caption{Pedestrian detection results on Caltech-USA test set. It is worth mentioning that all models are directly trained on Caltech-USA. We use $\rm MR^{-2}$ as the performance to compare each detectors. The best and the second best are highlighted in red and blue.}
\label{tab:caltech}
\end{table}

\subsection{Comparisons with State-of-the-art Methods}
\textbf{Result on CityPersons dataset.}
We compare our model with state-of-the-art pedestrian detection methods, including FRCNN~\cite{zhang2017citypersons}, RepLoss~\cite{wang2018repulsion}, OR-CNN~\cite{zhang2018occlusion}, ATT-part~\cite{zhang2018occluded}, Bi-Box~\cite{zhou2018bi}, MGAN~\cite{Pang_2019_ICCV}, ALFNet~\cite{Liu_2018_ECCV}, TLL~\cite{song2018small} and CSP~\cite{liu2018high} on CityPersons validation set. It is noticing that existing methods employ different detection framework and backbone, as well as different input scale, so we also list these components in Table~\ref{tab:cityperson} for fair comparison.

The performance results are summarized in Table~\ref{tab:cityperson}. It is evident that our model achieves best performance on \emph{Reasonable} subset, \eg outperforming the second best results by a margin of $0.6\%$. Comparing with CSP~\cite{liu2018high}, which is the current best region-based one-stage detector, we improve the $\rm MR^{-2}$ on \emph{Reasonable} subset from $11.0\%$ to $10.4\%$. It is worth mentioning that the extra anchor location branch is removable during inference, which makes the architecture of our detector no different than FRCNN~\cite{zhang2017citypersons} and RepLoss~\cite{wang2018repulsion}. We can observe that our methods surpasses these two models by $2.5\%$ / $1.2\%$ on \emph{Reasonable} subset and $3.6\%$ / $8.4\%$ on \emph{Heavy} subset. Models like OR-CNN~\cite{zhang2018occlusion}, ATT-part~\cite{zhang2018occluded}, Bi-Box~\cite{zhou2018bi}, MGAN~\cite{Pang_2019_ICCV} modify the network architecture in the second stage which lead to better performance under occlusion cases. Our model achieves $46.9\%$ on \emph{Heavy} subset, which is competitive with these models.

\noindent\textbf{Result on Caltech-USA dataset.}
We conduct extensively comparison with recent methods, including DeepParts~\cite{tian2015deep}, RPN+BF~\cite{zhang2016faster}, MS-CNN~\cite{cai2016unified}, SDS-RCNN~\cite{brazil2017illuminating}, ATT-part~\cite{zhang2018occluded}, RepLoss~\cite{wang2018repulsion}, Bi-Box~\cite{zhou2018bi}, and CSP~\cite{liu2018high}.

As shown in Table~\ref{tab:caltech}, our model achieves superior results comparing with most of the models and performs competitively with state-of-the-art method. Specifically, on \emph{Reasonable} subset, our model surpasses Bi-Box~\cite{zhou2018bi} by a margin of $2.7\%$ but sightly falls behind CSP~\cite{liu2018high} by $0.4\%$. Comparing with RepLoss~\cite{wang2018repulsion}, $\rm MR^{-2}$ on \emph{Heavy} and \emph{All} subsets reduce from $47.9\%$ to $45.5\%$ and $59.0\%$ to $57.0\%$ respectively.

\begin{table}[t]
\begin{center}
\setlength{\tabcolsep}{3.5pt}
\begin{tabular}{p{3.6cm}<{\centering}|p{2cm}<{\centering}|p{2cm}<{\centering}}
\toprule[1pt]
Model &{\em Reasonable} &{\em Heavy} \\
\hline
Baseline &12.7 &54.0 \\
+ IoULoss~\cite{yu2016unitbox} &12.4 &52.0 \\
+ RepLoss~\cite{wang2018repulsion} &11.6 &55.3 \\
+ AggLoss~\cite{zhang2018occlusion} &11.4 &52.6 \\
\hline
\hline
+ CouLoss (on RPN) &10.9 &53.0 \\
+ CouLoss (on RCNN) &11.0 &53.8 \\
+ CouLoss (only Att) &11.6 &53.0 \\
+ CouLoss (only Rep) &11.3 &52.5 \\
+ CouLoss &10.5 &51.9 \\
\bottomrule[1pt]
\end{tabular}
\end{center}
\caption{Comparison between CouLoss with other loss functions on CityPersons. ``on RPN'' and ``on RCNN'' represent using CouLoss only in RPN stage and Fast-RCNN stage, respectively. ``only Att'' and ``only Rep''on represent only including the work of Attractive Force and Repulsive Force, respectively.}
\label{tab:couloss}
\end{table}

\subsection{Ablation Study}
We carry out comprehensive ablation studies on CityPersons dataset to evaluate the contribution of each component in the proposed methods.

\begin{figure*}[t]
\centering
\includegraphics[width=1.0\linewidth]{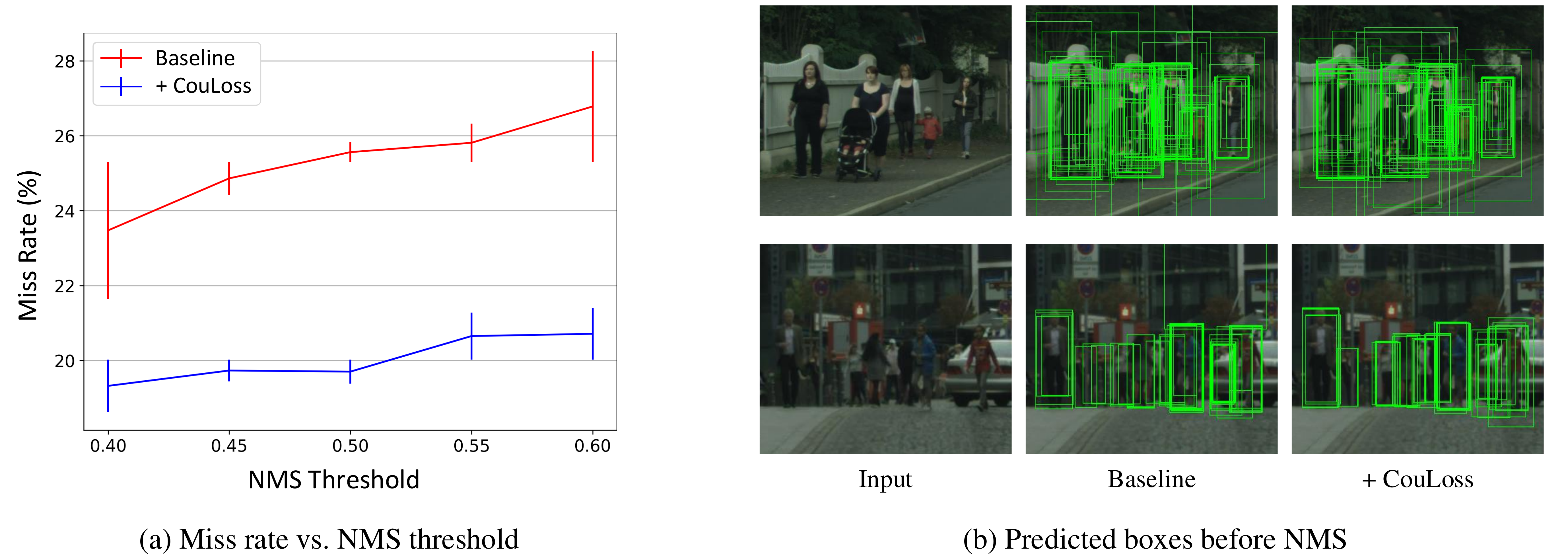}
\caption{(a) Comparison between baseline and CouLoss based on the miss rate across different nms thresholds. The curve of CouLoss is smoother than that of baseline, indicating that our method is less sensitive to nms threshold than baseline. The bar at each point represents the deviance from average value. (b) Visualization of predicted bounding boxes before nms. Compared with baseline, the predictions from CouLoss locate compactly around ground truths and there are fewer proposals lying in the overlaps between adjacent pedestrians }
\label{fig:coulossvis}
\end{figure*}

\noindent\textbf{Coulomb loss.} As shown in Table~\ref{tab:couloss}, we analyze the effectiveness of different components in the proposed CouLoss. The performance of using CouLoss on RPN stage is better than that of Fast-RCNN stage. The reason might be the former has larger scale of trainable bounding boxes than the latter. Only considering the \textbf{work} consumption of Repulsive Force achieves greater improvement than that of Attractive Force, this is because the original regression loss works as attractive term to some degree. Compared with baseline model, CouLoss achieves consistent lower $\rm MR^{-2}$ on the two listed subsets.

It is worth mentioning that our CouLoss is superior to the two state-of-the-art methods using AggLoss~\cite{zhang2018occlusion} and RepLoss~\cite{wang2018repulsion} which also serve as regularization terms on regression function. This proves that considering both attractive and repulsive constraints in the regularization term can further benefit region localization. When simply compare the attraction and repulsion terms among these methods, CouLoss still generates competitive performance. We also compare the results with IoULoss~\cite{zhang2018occlusion} since the forces in CouLoss are based on the form of IoULoss. The result shows that the promotion generated by our method is only marginally related to the using of IouLoss form.

Since CouLoss can pull proposals to their target ground truths and push them away from non-target ones, model becomes less sensitive to the NMS threshold. To demonstrate this point, we present the miss rate of CouLoss across various NMS threshold at $\rm FPPI=10^{-2}$.
In Figure ~\ref{fig:coulossvis}(a), model with CouLoss always produces lower miss rate than baseline. It is noteworthy that the curve of CouLoss is smoother than that of baseline, indicating that changing NMS threshold has less impact on CouLoss than on baseline. In addition, we also visualize the predicted bounding boxes before NMS in crowd scenes in Figure~\ref{fig:coulossvis}(b). Compared with baseline, the predictions of model trained with CouLoss locate compactly around ground truths and there are fewer proposals lying in the overlaps between adjacent pedestrians.

\begin{table}[t]
\begin{center}
\setlength{\tabcolsep}{4.5pt}
\begin{tabular}{c|p{2cm}<{\centering}|p{2cm}<{\centering}}
\toprule[1pt]
Model &{\em Reasonable} &{\em Heavy} \\
\hline
Baseline &12.7 / 0.22 &54.2 / 0.65 \\
+ Focal Loss~\cite{lin2017focal} &11.7 / 0.17 &53.8 / 0.60 \\
+ OHEM~\cite{shrivastava2016training} &11.4 / 0.15 &52.3 / 0.53 \\
\hline
\hline
+ anchor location branch &11.7 / 0.16 &53.2 / 0.54 \\
+ ALS (w/o IR) &11.9 / 0.17 &54.0 / 0.53 \\
+ ALS (w IR) &11.1 / 0.14 &51.2 / 0.48 \\
\bottomrule[1pt]
\end{tabular}
\end{center}
\caption{Comparison between ALS with other sampling methods in detection frameworks. The results are reported in the form of $\rm MR^{-2}/\rm FPPI$. The FPPI is calculated under $\rm MR=0.1$.}
\label{tab:anchorloc}
\end{table}

\begin{figure*}[t!]
\includegraphics[width=1.0\linewidth]{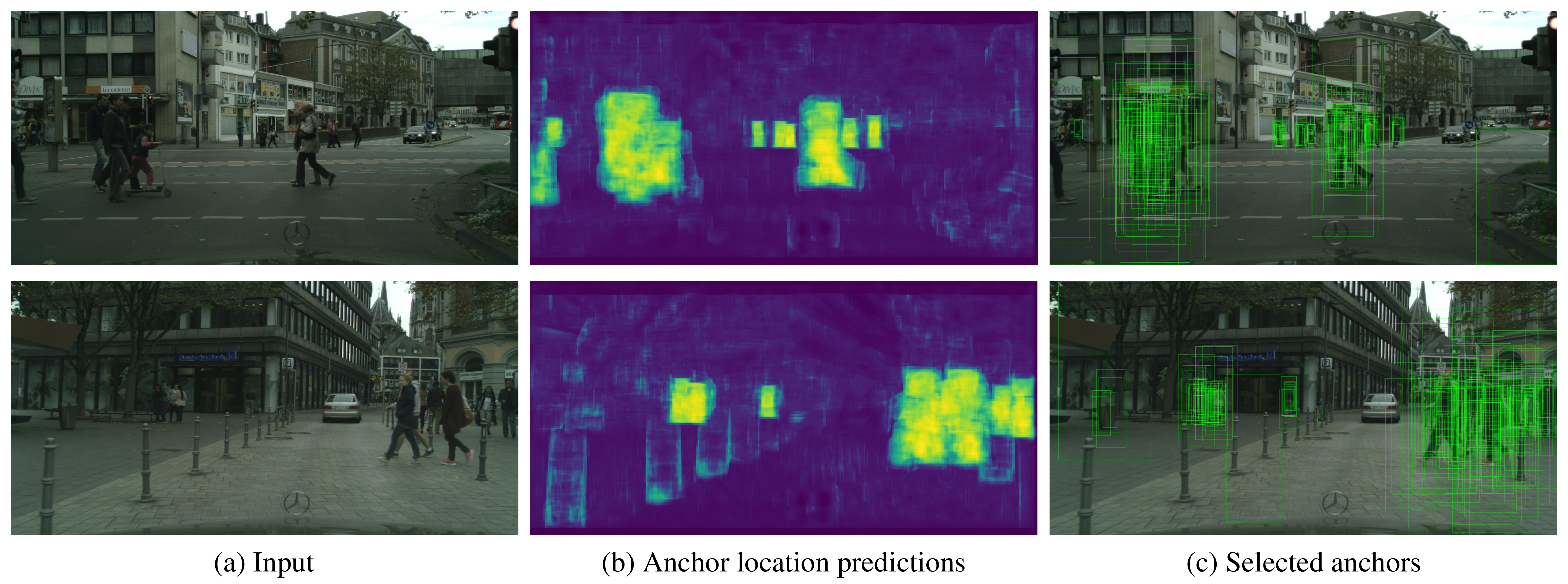}
\caption{The visualization of input images, generated probability maps, and selected anchors sampled by the proposed method.}
\label{fig:anchor}
\end{figure*}

\noindent\textbf{Anchor location selecting.} As listed in Table~\ref{tab:anchorloc}, ALS outperforms the two most widely used hard examples mining methods, \emph{i.e.} Focal Loss~\cite{lin2017focal} and OHEM~\cite{shrivastava2016training}, on both \emph{Reasonable} and \emph{Heavy} subsets. This proves that our method is more effective when handling human-like structures in pedestrian detection. To validate the design of ALS, we first set $\epsilon_a=0$ to bias the selecting process. The improvement on the forth row indicates that the extra anchor location branch is helpful in centering anchors around foregrounds. When further introducing the proposed selecting method in place of sliding window method, the model achieves $0.6\%$ and $2.0\%$ improvement on two subsets, respectively. This shows that the proposed ALS strategy can effectively mine the hard negatives by filtering out less-informative proposals from negative examples. Then we demonstrate the necessity of using ignored region (\emph{IR}) when training the anchor location branch. The results present that the model trained without \emph{IR} performs consistently worse on both subsets. This is mainly because the model miss-labels proposals as positive examples when they are largely occupied by non-visible parts, as discussed in Section~\ref{sec:anchorloc}. Since the proposed method is specially designed for false positive cases which can't be clearly reflected on $\rm MR^{-2}$, we use FPPI to evaluate the model and observe consistent lower score on all subsets.

Results mentioned above are also supported by the visualization results shown in Figure~\ref{fig:anchor}, where we present input images, generated probability maps, and selected anchors sampled by the proposed method. It can be seen that the probability maps in Figure~\ref{fig:anchor}(b) are highly correlated with human-shape structures, which leads the selected anchors to concentrate more on these objects as shown in Figures~\ref{fig:anchor}(c).


\section{Extension: Results on PASCAL VOC}
In this section, we extend the application of our proposed methods to reveal its universality. The modifications are applied on general object detection application which also suffers from occluded scenes and hard negative examples.

Our experiments are performed on PASCAL VOC dataset~\cite{everingham2010pascal} which is a common benchmark for general objection detection. We employ Faster R-CNN with ResNet-101~\cite{he2016deep} as the backbone for baseline detector. The model is trained on the training and validation sets of PASCAL VOC 2007 and PASCAL VOC 2012 without any bells and whistles, and is tested on the testing set of PASCAL VOC 2007. To evaluate high quality detection results from our methods, we use the COCO metrics for evaluation\footnote[3]{The annotations of PASCAL VOC are transformed to COCO format and COCO toolbox is used for evaluation.}. The results in Table \ref{tab:voc} show that the proposed methods have significant improvement on general object detection task, especially under high IoU threshold, which demonstrates the universality of the proposed methods.

\begin{table}[t]
\begin{center}
\setlength{\tabcolsep}{5.5pt}
\begin{tabular}{c|p{1.0cm}<{\centering}p{1.0cm}<{\centering}p{1.0cm}<{\centering}}
\toprule[1pt]
Method &$AP$ &$AP_{50}$ &$AP_{75}$ \\
\hline
Faster R-CNN & 49.2 & 77.2 & 53.8 \\
+ CouLoss & +0.8 & +0.2 & +1.0 \\
+ ALS & +0.7 & +0.2 & +1.1 \\
+ All & +2.3 & +0.5 & +3.7 \\
\bottomrule[1pt]
\end{tabular}
\end{center}
\caption{General object detection results on PASCAL VOC 2007 test set. Note that these results are evaluated under COCO metrics which are different from the original VOC metrics.}
\label{tab:voc}
\end{table}


\section{Conclusion}
In this paper, we systematically analyze the impact of crowd occlusion and hard negatives on the general region-based detection framework, \emph{i.e.} the deterioration of bounding box localization and classification. We then put forward two specific approaches to tackle these two barriers to pedestrian detection in wild. For the former, we design a new loss function, termed as CouLoss, to regulate the process of bounding box regression. Specifically, we build a \textbf{physics} framework to unify the process of attraction and repulsion, which can pull proposals towards their target ground truths and push proposals away from non-target ones respectively. For the latter, an efficient semantic-driven strategy for selecting anchor locations is introduced, which can sample human-like structures as informative negative examples during training phase for classification refinement. It is worth mentioning that both methods don't increase any computational cost during inference phase.

Our methods can be trained in an end-to-end fashion and achieves competitive performance on two widely adopted benchmarking datasets, \emph{i.e.} Caltech-USA and CityPersons. Detailed ablation experiments have demonstrated the effectiveness of each approach respectively. More importantly, the promising preliminary results on PASCAL VOC show that our methods could also be adopted towards other appearance-based object detection tasks.

\clearpage

{\small
\bibliographystyle{ieee_fullname}
\bibliography{egbib}

\begin{thebibliography}{10}\itemsep=-1pt

\bibitem{benenson2014ten}
Rodrigo Benenson, Mohamed Omran, Jan Hosang, and Bernt Schiele.
\newblock Ten years of pedestrian detection, what have we learned?
\newblock In {\em European Conference on Computer Vision}, pages 613--627.
  Springer, 2014.

\bibitem{Bodla_2017_ICCV}
Navaneeth Bodla, Bharat Singh, Rama Chellappa, and Larry~S. Davis.
\newblock Soft-nms -- improving object detection with one line of code.
\newblock In {\em The IEEE International Conference on Computer Vision (ICCV)},
  Oct 2017.

\bibitem{brazil2017illuminating}
Garrick Brazil, Xi Yin, and Xiaoming Liu.
\newblock Illuminating pedestrians via simultaneous detection \& segmentation.
\newblock {\em arXiv preprint arXiv:1706.08564}, 2017.

\bibitem{cai2016unified}
Zhaowei Cai, Quanfu Fan, Rogerio~S Feris, and Nuno Vasconcelos.
\newblock A unified multi-scale deep convolutional neural network for fast
  object detection.
\newblock In {\em european conference on computer vision}, pages 354--370.
  Springer, 2016.

\bibitem{cao2019prime}
Yuhang Cao, Kai Chen, Chen~Change Loy, and Dahua Lin.
\newblock Prime sample attention in object detection.
\newblock {\em arXiv preprint arXiv:1904.04821}, 2019.

\bibitem{mmdetection}
Kai Chen, Jiaqi Wang, Jiangmiao Pang, Yuhang Cao, Yu Xiong, Xiaoxiao Li,
  Shuyang Sun, Wansen Feng, Ziwei Liu, Jiarui Xu, Zheng Zhang, Dazhi Cheng,
  Chenchen Zhu, Tianheng Cheng, Qijie Zhao, Buyu Li, Xin Lu, Rui Zhu, Yue Wu,
  Jifeng Dai, Jingdong Wang, Jianping Shi, Wanli Ouyang, Chen~Change Loy, and
  Dahua Lin.
\newblock {MMDetection}: Open mmlab detection toolbox and benchmark.
\newblock {\em arXiv preprint arXiv:1906.07155}, 2019.

\bibitem{cordts2016cityscapes}
Marius Cordts, Mohamed Omran, Sebastian Ramos, Timo Rehfeld, Markus Enzweiler,
  Rodrigo Benenson, Uwe Franke, Stefan Roth, and Bernt Schiele.
\newblock The cityscapes dataset for semantic urban scene understanding.
\newblock In {\em Proceedings of the IEEE conference on computer vision and
  pattern recognition}, pages 3213--3223, 2016.

\bibitem{dalal2005histograms}
Navneet Dalal and Bill Triggs.
\newblock Histograms of oriented gradients for human detection.
\newblock In {\em Computer Vision and Pattern Recognition, 2005. CVPR 2005.
  IEEE Computer Society Conference on}, volume~1, pages 886--893. IEEE, 2005.

\bibitem{dollar2009integral}
Piotr Doll{\'a}r, Zhuowen Tu, Pietro Perona, and Serge Belongie.
\newblock Integral channel features.
\newblock 2009.

\bibitem{dollar2009pedestrian}
Piotr Doll{\'a}r, Christian Wojek, Bernt Schiele, and Pietro Perona.
\newblock Pedestrian detection: A benchmark.
\newblock In {\em Computer Vision and Pattern Recognition, 2009. CVPR 2009.
  IEEE Conference on}, pages 304--311. IEEE, 2009.

\bibitem{du2017fused}
Xianzhi Du, Mostafa El-Khamy, Jungwon Lee, and Larry Davis.
\newblock Fused dnn: A deep neural network fusion approach to fast and robust
  pedestrian detection.
\newblock In {\em Applications of Computer Vision (WACV), 2017 IEEE Winter
  Conference on}, pages 953--961. IEEE, 2017.

\bibitem{ess2008mobile}
Andreas Ess, Bastian Leibe, Konrad Schindler, and Luc Van~Gool.
\newblock A mobile vision system for robust multi-person tracking.
\newblock In {\em Computer Vision and Pattern Recognition, 2008. CVPR 2008.
  IEEE Conference on}, pages 1--8. IEEE, 2008.

\bibitem{everingham2010pascal}
Mark Everingham, Luc Van~Gool, Christopher~KI Williams, John Winn, and Andrew
  Zisserman.
\newblock The pascal visual object classes (voc) challenge.
\newblock {\em International journal of computer vision}, 88(2):303--338, 2010.

\bibitem{girshick2015fast}
Ross Girshick.
\newblock Fast r-cnn.
\newblock In {\em Proceedings of the IEEE international conference on computer
  vision}, pages 1440--1448, 2015.

\bibitem{girshick2014rich}
Ross Girshick, Jeff Donahue, Trevor Darrell, and Jitendra Malik.
\newblock Rich feature hierarchies for accurate object detection and semantic
  segmentation.
\newblock In {\em Proceedings of the IEEE conference on computer vision and
  pattern recognition}, pages 580--587, 2014.

\bibitem{halliday2013fundamentals}
David Halliday, Robert Resnick, and Jearl Walker.
\newblock {\em Fundamentals of physics}.
\newblock John Wiley \& Sons, 2013.

\bibitem{he2016deep}
Kaiming He, Xiangyu Zhang, Shaoqing Ren, and Jian Sun.
\newblock Deep residual learning for image recognition.
\newblock In {\em Proceedings of the IEEE conference on computer vision and
  pattern recognition}, pages 770--778, 2016.

\bibitem{huang2015densebox}
Lichao Huang, Yi Yang, Yafeng Deng, and Yinan Yu.
\newblock Densebox: Unifying landmark localization with end to end object
  detection.
\newblock {\em arXiv preprint arXiv:1509.04874}, 2015.

\bibitem{li2019gradient}
Buyu Li, Yu Liu, and Xiaogang Wang.
\newblock Gradient harmonized single-stage detector.
\newblock In {\em Proceedings of the AAAI Conference on Artificial
  Intelligence}, volume~33, pages 8577--8584, 2019.

\bibitem{lin2018graininess}
Chunze Lin, Jiwen Lu, Gang Wang, and Jie Zhou.
\newblock Graininess-aware deep feature learning for pedestrian detection.
\newblock In {\em Proceedings of the European Conference on Computer Vision
  (ECCV)}, pages 732--747, 2018.

\bibitem{Lin_2017_ICCV}
Tsung-Yi Lin, Priya Goyal, Ross Girshick, Kaiming He, and Piotr Dollar.
\newblock Focal loss for dense object detection.
\newblock In {\em The IEEE International Conference on Computer Vision (ICCV)},
  Oct 2017.

\bibitem{lin2017focal}
Tsung-Yi Lin, Priya Goyal, Ross Girshick, Kaiming He, and Piotr Doll{\'a}r.
\newblock Focal loss for dense object detection.
\newblock In {\em Proceedings of the IEEE international conference on computer
  vision}, pages 2980--2988, 2017.

\bibitem{Liu_2019_CVPR}
Songtao Liu, Di Huang, and Yunhong Wang.
\newblock Adaptive nms: Refining pedestrian detection in a crowd.
\newblock In {\em The IEEE Conference on Computer Vision and Pattern
  Recognition (CVPR)}, June 2019.

\bibitem{Liu_2018_ECCV}
Wei Liu, Shengcai Liao, Weidong Hu, Xuezhi Liang, and Xiao Chen.
\newblock Learning efficient single-stage pedestrian detectors by asymptotic
  localization fitting.
\newblock In {\em The European Conference on Computer Vision (ECCV)}, September
  2018.

\bibitem{mao2017can}
Jiayuan Mao, Tete Xiao, Yuning Jiang, and Zhimin Cao.
\newblock What can help pedestrian detection?
\newblock In {\em 2017 IEEE Conference on Computer Vision and Pattern
  Recognition (CVPR)}, pages 6034--6043. IEEE, 2017.

\bibitem{noh2018improving}
Junhyug Noh, Soochan Lee, Beomsu Kim, and Gunhee Kim.
\newblock Improving occlusion and hard negative handling for single-stage
  pedestrian detectors.
\newblock In {\em Proceedings of the IEEE Conference on Computer Vision and
  Pattern Recognition}, pages 966--974, 2018.

\bibitem{pang2019libra}
Jiangmiao Pang, Kai Chen, Jianping Shi, Huajun Feng, Wanli Ouyang, and Dahua
  Lin.
\newblock Libra r-cnn: Towards balanced learning for object detection.
\newblock In {\em Proceedings of the IEEE Conference on Computer Vision and
  Pattern Recognition}, pages 821--830, 2019.

\bibitem{Pang_2019_ICCV}
Yanwei Pang, Jin Xie, Muhammad~Haris Khan, Rao~Muhammad Anwer, Fahad~Shahbaz
  Khan, and Ling Shao.
\newblock Mask-guided attention network for occluded pedestrian detection.
\newblock In {\em The IEEE International Conference on Computer Vision (ICCV)},
  October 2019.

\bibitem{ren2015faster}
Shaoqing Ren, Kaiming He, Ross Girshick, and Jian Sun.
\newblock Faster r-cnn: Towards real-time object detection with region proposal
  networks.
\newblock In {\em Advances in neural information processing systems}, pages
  91--99, 2015.

\bibitem{shrivastava2016training}
Abhinav Shrivastava, Abhinav Gupta, and Ross Girshick.
\newblock Training region-based object detectors with online hard example
  mining.
\newblock In {\em Proceedings of the IEEE Conference on Computer Vision and
  Pattern Recognition}, pages 761--769, 2016.

\bibitem{simonyan2014very}
Karen Simonyan and Andrew Zisserman.
\newblock Very deep convolutional networks for large-scale image recognition.
\newblock {\em arXiv preprint arXiv:1409.1556}, 2014.

\bibitem{Song_2018_ECCV}
Tao Song, Leiyu Sun, Di Xie, Haiming Sun, and Shiliang Pu.
\newblock Small-scale pedestrian detection based on topological line
  localization and temporal feature aggregation.
\newblock In {\em The European Conference on Computer Vision (ECCV)}, September
  2018.

\bibitem{song2018small}
Tao Song, Leiyu Sun, Di Xie, Haiming Sun, and Shiliang Pu.
\newblock Small-scale pedestrian detection based on topological line
  localization and temporal feature aggregation.
\newblock In {\em Proceedings of the European Conference on Computer Vision
  (ECCV)}, pages 536--551, 2018.

\bibitem{tian2015deep}
Yonglong Tian, Ping Luo, Xiaogang Wang, and Xiaoou Tang.
\newblock Deep learning strong parts for pedestrian detection.
\newblock In {\em Proceedings of the IEEE international conference on computer
  vision}, pages 1904--1912, 2015.

\bibitem{Tian_2019_ICCV}
Zhi Tian, Chunhua Shen, Hao Chen, and Tong He.
\newblock Fcos: Fully convolutional one-stage object detection.
\newblock In {\em The IEEE International Conference on Computer Vision (ICCV)},
  October 2019.

\bibitem{wang2019region}
Jiaqi Wang, Kai Chen, Shuo Yang, Chen~Change Loy, and Dahua Lin.
\newblock Region proposal by guided anchoring.
\newblock {\em arXiv preprint arXiv:1901.03278}, 2019.

\bibitem{wang2018repulsion}
Xinlong Wang, Tete Xiao, Yuning Jiang, Shuai Shao, Jian Sun, and Chunhua Shen.
\newblock Repulsion loss: detecting pedestrians in a crowd.
\newblock In {\em Proceedings of the IEEE Conference on Computer Vision and
  Pattern Recognition}, pages 7774--7783, 2018.

\bibitem{liu2018high}
Weiqiang Ren Weidong Hu Yinan~Yu Wei~Liu, Shengcai~Liao.
\newblock High-level semantic feature detection: A new perspective for
  pedestrian detection.
\newblock In {\em IEEE Conference on Computer Vision and Pattern Recognition
  (CVPR)}, 2019.

\bibitem{yu2016unitbox}
Jiahui Yu, Yuning Jiang, Zhangyang Wang, Zhimin Cao, and Thomas Huang.
\newblock Unitbox: An advanced object detection network.
\newblock In {\em Proceedings of the 2016 ACM on Multimedia Conference}, pages
  516--520. ACM, 2016.

\bibitem{zhang2016faster}
Liliang Zhang, Liang Lin, Xiaodan Liang, and Kaiming He.
\newblock Is faster r-cnn doing well for pedestrian detection?
\newblock In {\em European Conference on Computer Vision}, pages 443--457.
  Springer, 2016.

\bibitem{zhang2016far}
Shanshan Zhang, Rodrigo Benenson, Mohamed Omran, Jan Hosang, and Bernt Schiele.
\newblock How far are we from solving pedestrian detection?
\newblock In {\em Proceedings of the IEEE Conference on Computer Vision and
  Pattern Recognition}, pages 1259--1267, 2016.

\bibitem{zhang2017citypersons}
Shanshan Zhang, Rodrigo Benenson, and Bernt Schiele.
\newblock Citypersons: A diverse dataset for pedestrian detection.
\newblock In {\em The IEEE Conference on Computer Vision and Pattern
  Recognition (CVPR)}, volume~1, page~3, 2017.

\bibitem{zhang2018occlusion}
Shifeng Zhang, Longyin Wen, Xiao Bian, Zhen Lei, and Stan~Z Li.
\newblock Occlusion-aware r-cnn: detecting pedestrians in a crowd.
\newblock In {\em Proceedings of the European Conference on Computer Vision
  (ECCV)}, pages 637--653, 2018.

\bibitem{zhang2018occluded}
Shanshan Zhang, Jian Yang, and Bernt Schiele.
\newblock Occluded pedestrian detection through guided attention in cnns.
\newblock In {\em Proceedings of the IEEE Conference on Computer Vision and
  Pattern Recognition}, pages 6995--7003, 2018.

\bibitem{zhou2018bi}
Chunluan Zhou and Junsong Yuan.
\newblock Bi-box regression for pedestrian detection and occlusion estimation.
\newblock In {\em Proceedings of the European Conference on Computer Vision
  (ECCV)}, pages 135--151, 2018.

\end{thebibliography}
}

\end{document}